\documentclass{article}
\usepackage{spconf,amsmath,graphicx}

\usepackage[utf8]{inputenc}
\usepackage[english]{babel}
\usepackage{color}
\usepackage{url}
\usepackage{caption}
\usepackage{multirow}
\usepackage{multicol}
\usepackage{algorithm2e}
\usepackage{todonotes}

\newcommand{\en}{\enspace} 
 
\title{Ink removal from histopathology whole slide images by combining classification, detection and image generation models}
\name{Sharib Ali$^{\star}$ \en Nasullah Khalid Alham$^{\star,\dagger}$ \en Clare Verrill$^{\dagger}$ \en Jens Rittscher$^{\star}$}
\address{${\star}$Institute of Biomedical Engineering, Department of Engineering Science, \\
University of Oxford, Oxford\\
${\dagger}$Nuffield Department of Surgical Sciences and Oxford NIHR Biomedical Research Centre (BRC), \\University of Oxford, John Radcliffe Hospital, Oxford}
\begin{document}
%
\maketitle
\begin{abstract}
\noindent{Histopathology} slides are routinely marked by pathologists using permanent ink markers that should not be removed as they form part of the medical record. Often tumour regions are marked up for the purpose of highlighting features or other downstream processing such an gene sequencing. Once digitised there is no established method for removing this information from the whole slide images limiting its usability in research and study.~Removal of marker ink from these high-resolution whole slide images is non-trivial and complex problem as they contaminate different regions and in an inconsistent manner. We propose an efficient pipeline using convolution neural networks that results in ink-free images without compromising information and image resolution.~Our pipeline includes a sequential classical convolution neural network for accurate classification of contaminated image tiles, a fast region detector and a domain adaptive cycle consistent adversarial generative model for restoration of foreground pixels.~Both quantitative and qualitative results on four different whole slide images show that our approach yields visually coherent ink-free whole slide images.
\end{abstract}
\section{Introduction}

Histopathology slides are routinely marked by pathologists using permanent ink makers before imaging that should not usually be removed as these markings form part of the medical record.~Once the slides are digitised, these markings contaminate the images.~Such images are not advisable to be used for research and educational purposes as these markings often contain information which could bias results of a study. Slides are for example marked with “EPE” referring to extra-prostatic extension could be a visual prompt for a pathologist.

In addition,  Digital images contaminated with ink markers also obstruct further analysis as many regions could be submerged into a dark ink blots appearing in digital images and absence of clear structure of the underlying cells. 
To enable the use of these slides we propose a method for removing these ink marks and use domain adaptive cycle consistent adversarial generative model for restoration of foreground pixels. This work is based on the assumption that the ink marks do not obscure any diagnostically relevant information. 

\begin{figure}[t!]
\centering
\includegraphics[trim=0.25cm 0.25cm 0.0cm 0.1cm, clip=true,scale=0.18]{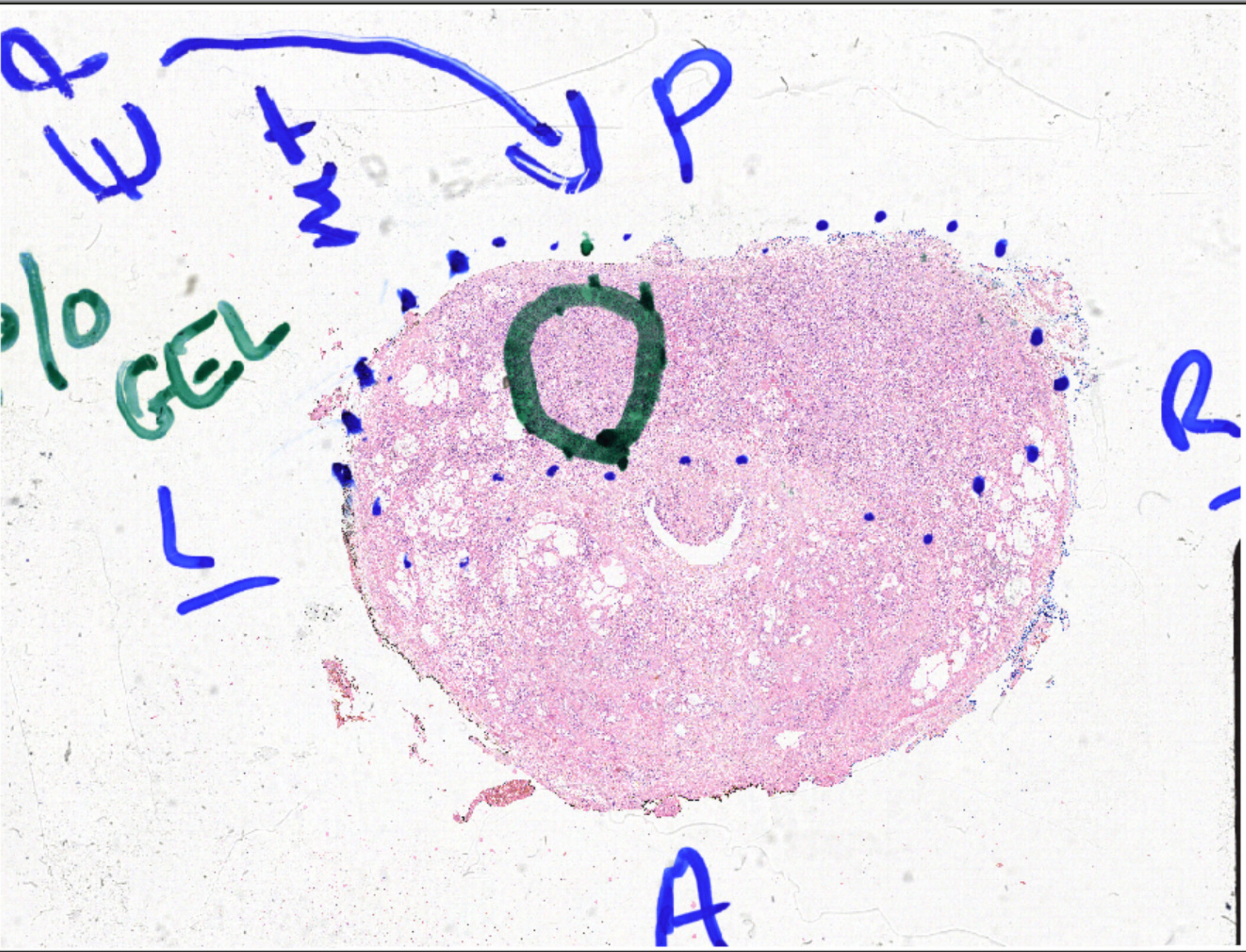}
\includegraphics[trim=0.25cm 0.25cm 0.5cm 0.8cm, clip=true,scale=0.2]{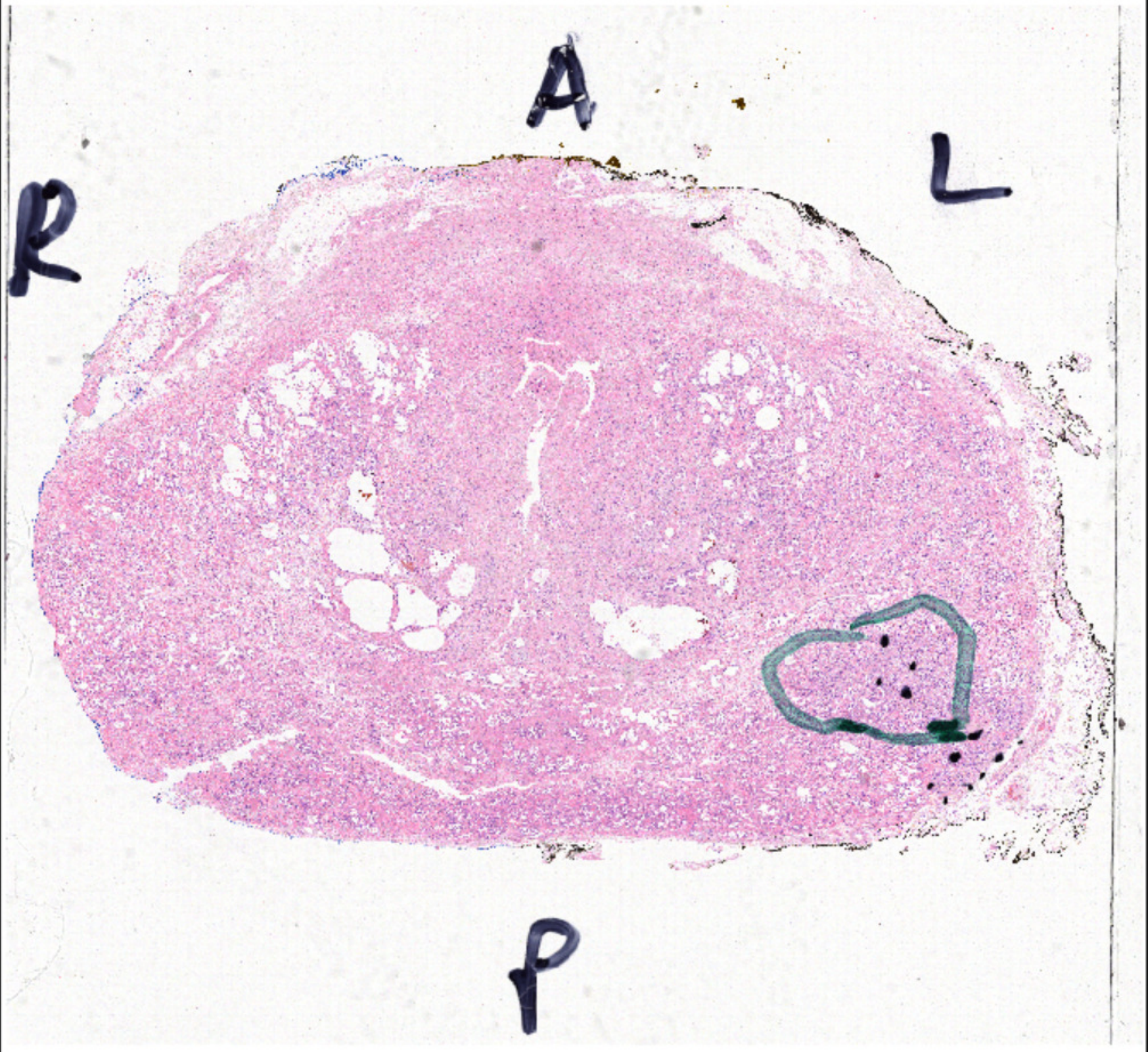}
\caption{{\bf Marked histology slides.} \it Shown are two whole slide images which have been marked by pathologists. For example, the circle on the left indicates an area of tumour that should be used for subsequent sequencing. Our goal is to remove such ink marks from digitised images so that these slides can be used for teaching and computational analysis. {\label{fig:inkcontamination}}}
\end{figure}
Ink (referred to \textit{``marker ink''} in this paper) markings can of course be made by pens of different colours. These markings can be on the tissue or outside around the tissue or both. Also, they are often composed of inconsistent patterns comprising of various elements like letters, circles, arrows, numbers or dots or a combination of them (see Fig.~\ref{fig:inkcontamination}). Ink removal from corrupted histopathogy images is very challenging as these tissue samples also possess inter- and intra-texture variabilities (e.g. due to inconsistency in staining).

Established methods that are capable of compensating for staining variations cannot be used to remove ink marks. 
Macenko {\it et al.} \cite{Macenko09:ISBI} convert RGB colours into corresponding optical density values to compensate for variations in hematoxylin and eosin staining. Niethammer and colleagues~ \cite{Niethammer10:MLMI} propose a colour correction model based on a similar approach but an improved estimation technique. A look-up table (LUT) based on the dye concentration absorbed by the sample was built in~\cite{Bautista2015} to correct for staining inconsistencies. 
Such colour transformations will not be sufficient to remove high opacity signals such as permanent ink markings. Given the texture variation of the original tissue image static transformations like those stored in look up tables cannot be used to restore the original image. 

\begin{figure}[t]
    \centering
    \begin{minipage}[b]{0.35\linewidth}
       \includegraphics[scale=0.23]{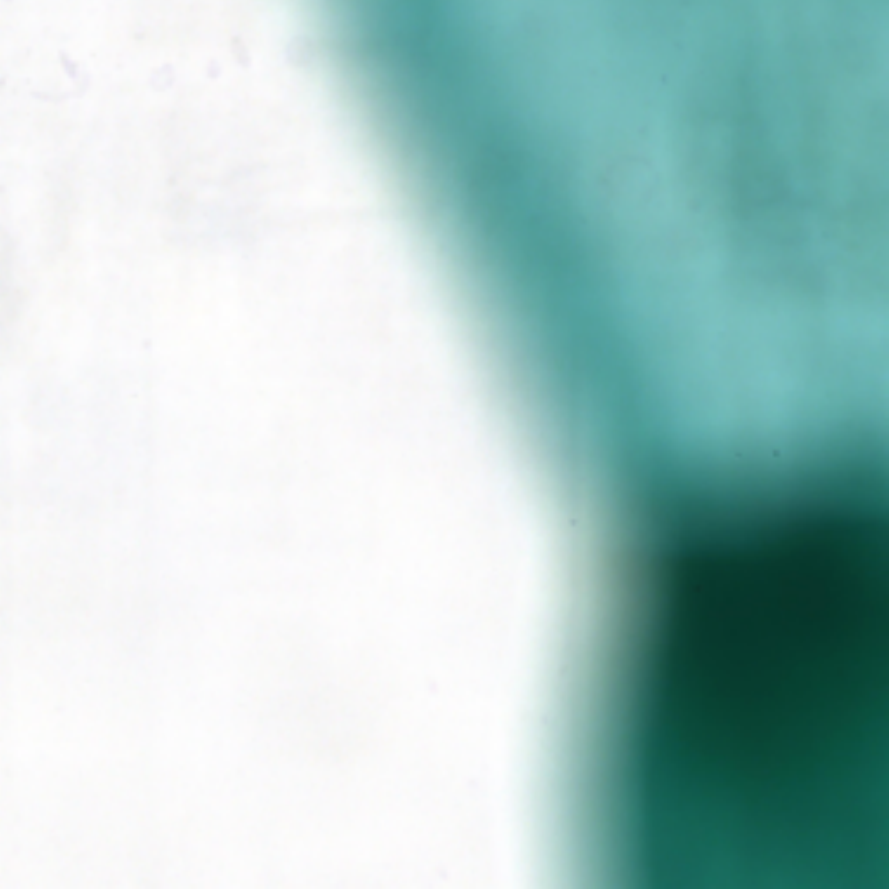}
    \end{minipage}
    \begin{minipage}[b]{0.35\linewidth}
    \includegraphics[scale=0.23]{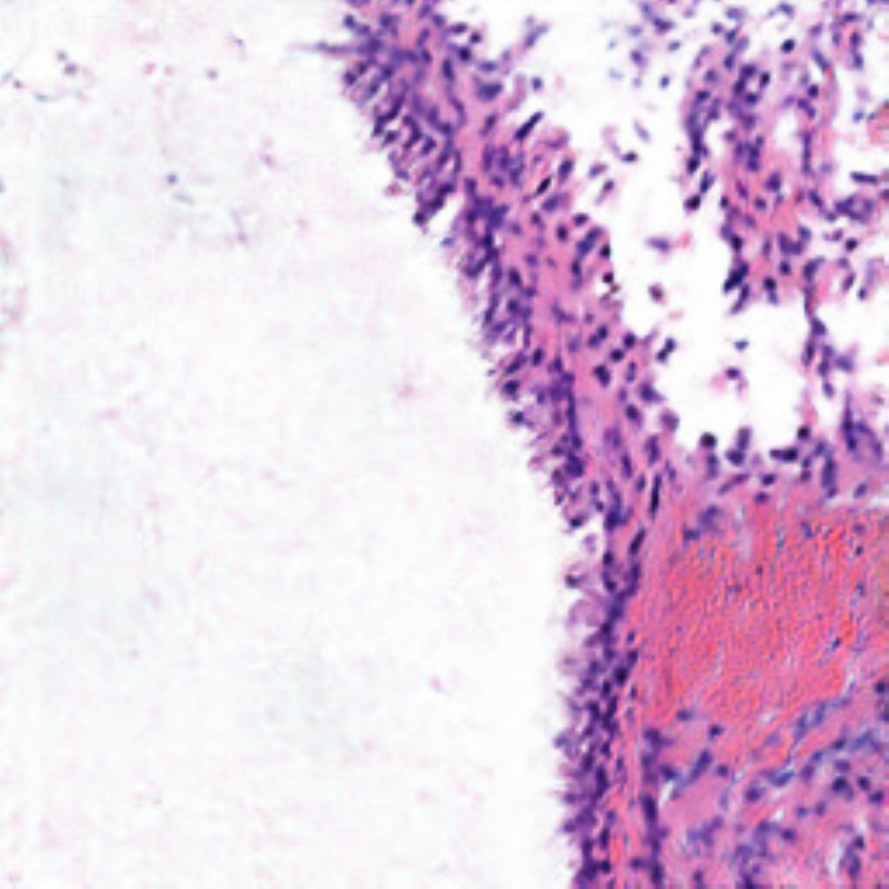}
     \end{minipage}
    \caption{{\bf Unrealistic hallucinations.} \it The naive application of generative models using CycleGAN~\cite{CycleGAN2017} can lead to undesired results.}
    \label{fig:GANS_halucination}
\end{figure}

Recent developments in deep learning based adversarial networks~\cite{Goodfellow:NIPS14} allows the modelling of complex data distributions that might not be captured by classical approaches. When compared to classical stain normalisation approaches Cycle consistent Generative Adversarial Networks (CycleGANs~\cite{CycleGAN2017}) can lead to improved results~\cite{Shaban2018StainGANSS}. While these models are very powerful, the example shown in 
Figure~\ref{fig:GANS_halucination} illustrates that any naive application of such models can lead to implausible results. In order to minimize such unrealistic hallucinations, we propose a fully automatic CNN-based approach that includes: 1) separating background ink corruption from foreground ink corruption, 2) identifying regions in image tiles with ink, and sparse or compactly arranged cells or cytoplasms and 3) domain learning, separately for sparse and densely arranged cell or cytoplasmic distributions. Finally, we have evaluated our approach on four different whole slide images contaminated with marker inks. 
%
%
\vspace{-0.25cm}
\section{Method}
\begin{figure}
    \centering
    \includegraphics[scale=0.25]{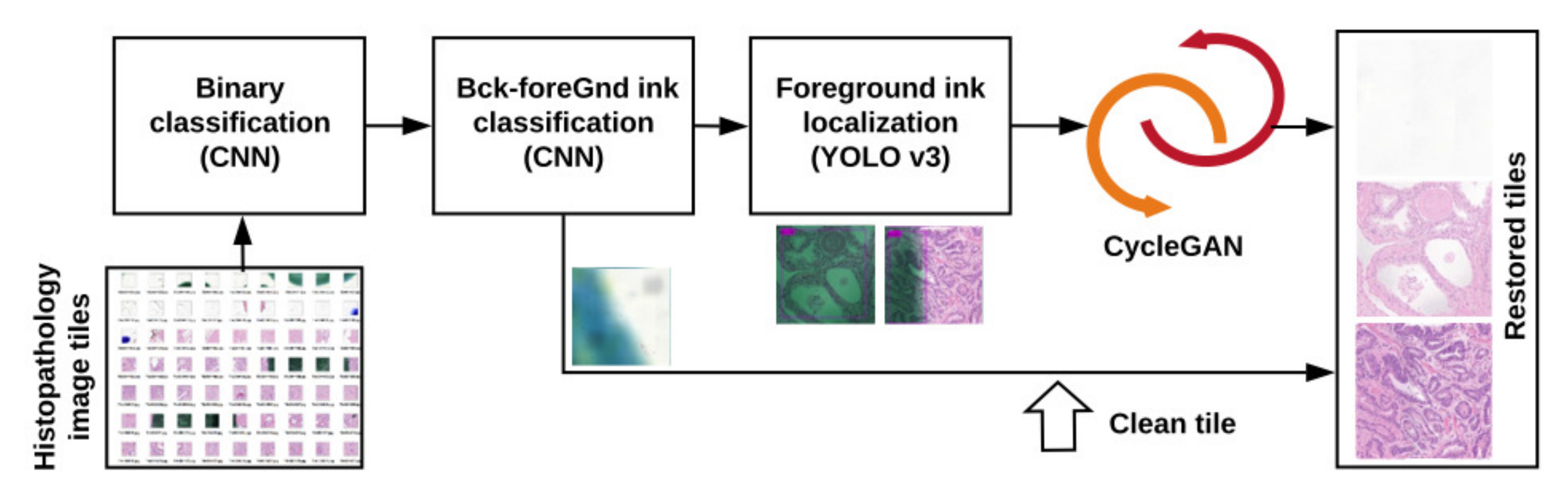}
    \caption{{\bf Proposed appraoch} {\it Rather than applying a global transformation model, CNN based binary detectors are used to delineate the ink markings. Subsequently, generative models are used to replace the missing information. First binary classification block outputs ink contaminated image tiles which is then fed to another sequential CNN classifier that separates image tiles with only background ink from foreground ink. Identified image tiles having only background ink are replaced with clean image tile (white background in our case) while the foreground ink contaminated tile goes to the foreground ink and cluster localization block.}}
    \label{fig:sequenctial_CNN}
\end{figure}

We propose a fully automatic convolution neural network based approach~(see Fig.~\ref{fig:sequenctial_CNN}) for the \textit{classification and restoration} of whole slide images containing ink marks in different colours. Our model includes: 1) Sequential CNN based architecture for binary classification, 2) Yolov3~\cite{redmon2018yolov3} for detection of bounding boxes for ink corrupted and cell cluster regions, and 3) CycleGAN~\cite{CycleGAN2017} for domain learning. Each of these process are described below. The code for the presented pipeline with trained weights and sample images used in this paper has been made publicly available \footnote{https://github.com/sharibox/histopathology-inkRemoval}. This tool is only for research use and is not validated for use in a diagnostic setting.
\subsection{Sequential CNN for classification of contaminated image tiles}
\begin{figure}[t!]
    \centering
    \includegraphics[scale=0.27]{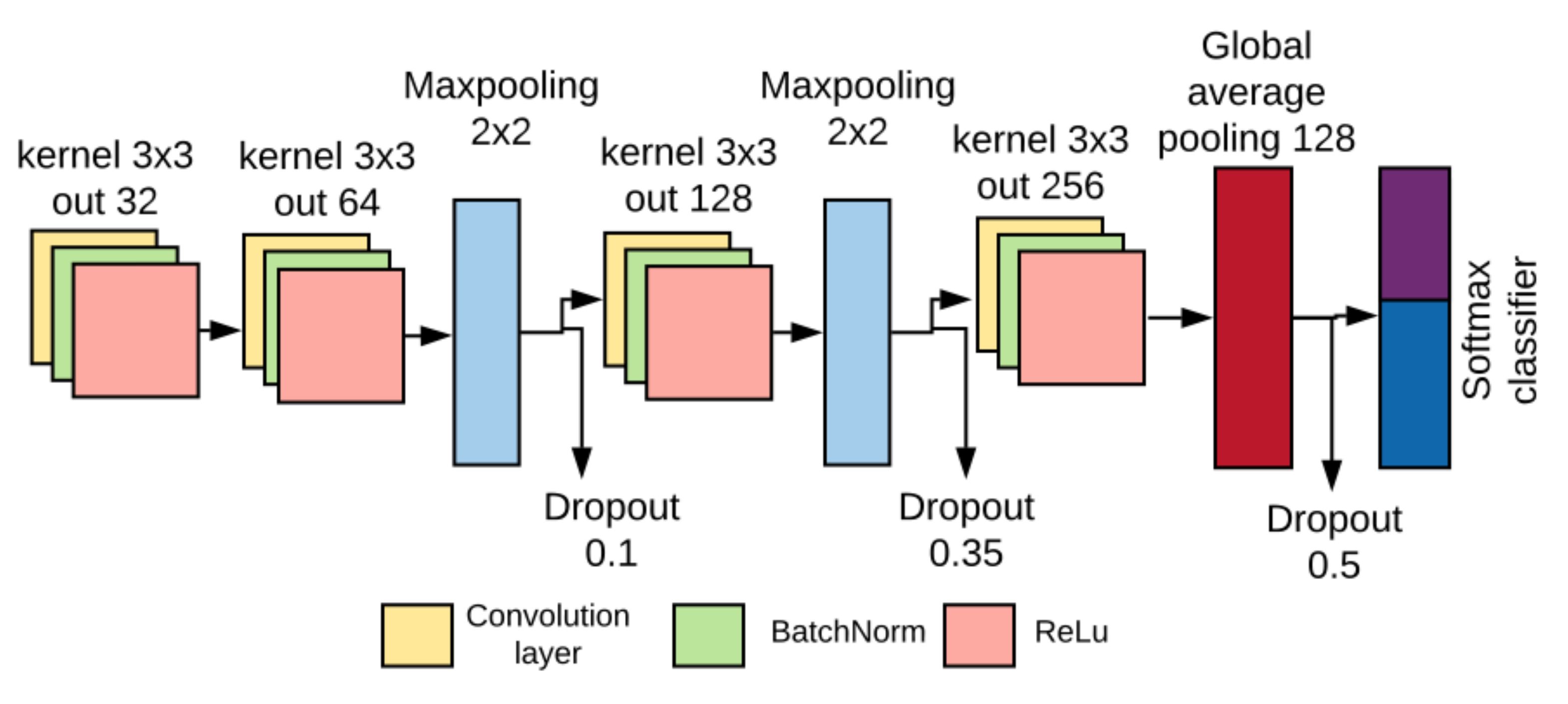}
    \caption{{\bf Binary classifier.} {\it The identical CNN architecture is used to classify if an entire tile or a pixel location contains ink marks (see Figure \ref{fig:sequenctial_CNN}). The model has 433K trainable parameters and permits rapid training and near real-time execution.}}
    \label{fig:sequenctial_CNN-2}
\end{figure}

We propose to use a sequential convolution neural network architecture shown in Figure~\ref{fig:sequenctial_CNN-2} for our two step binary classification: 1) classification of image tiles contaminated with ink from non-ink ones, and 2) classification of background contamination from the foreground. We have used cross entropy as loss function and an Adam optimizer with learning rate of 0.00001. The used model has only 433K trainable parameters. A total of $1000$ samples were utilized to train the tile classification model and $1746$ samples were used for pixel level background/foreground classification. In each case, we have used 40\% samples for validation and the tiles incorporate from 4 whole slide images used for training only.
\subsection{Precise localization of ink  and cluster in the foreground}{\label{sec:detector}}
\begin{figure}[t!]
\centering
\begin{minipage}[b]{0.32\linewidth}
\includegraphics[scale=0.4]{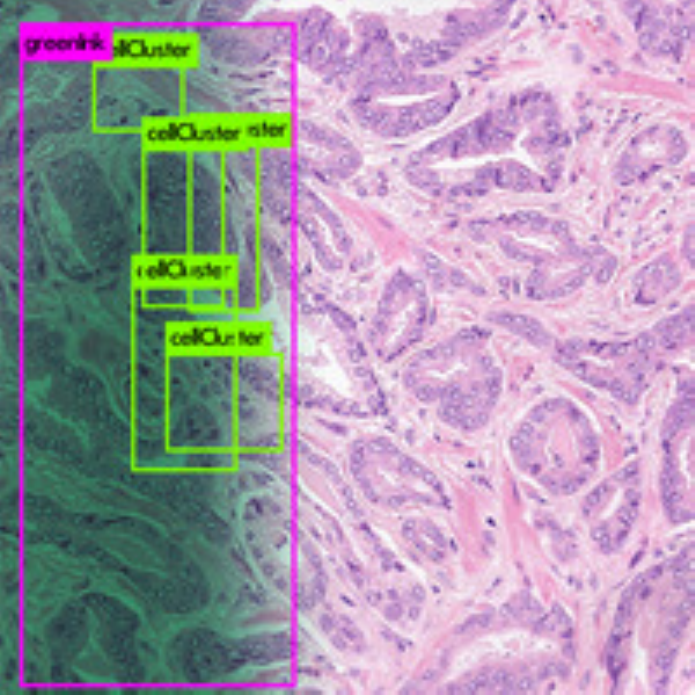}
\end{minipage}
\begin{minipage}[b]{0.32\linewidth}
\includegraphics[scale=0.4]{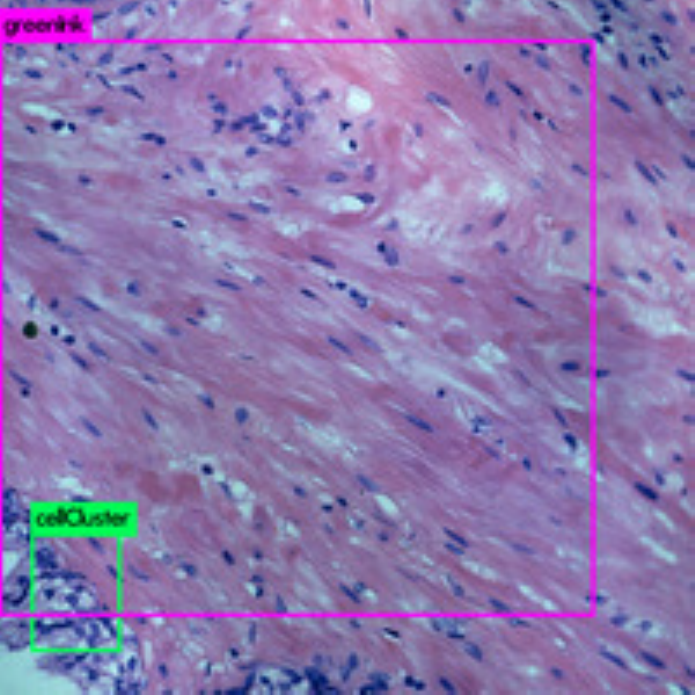}
\end{minipage}
\begin{minipage}[b]{0.32\linewidth}
\includegraphics[scale=0.4]{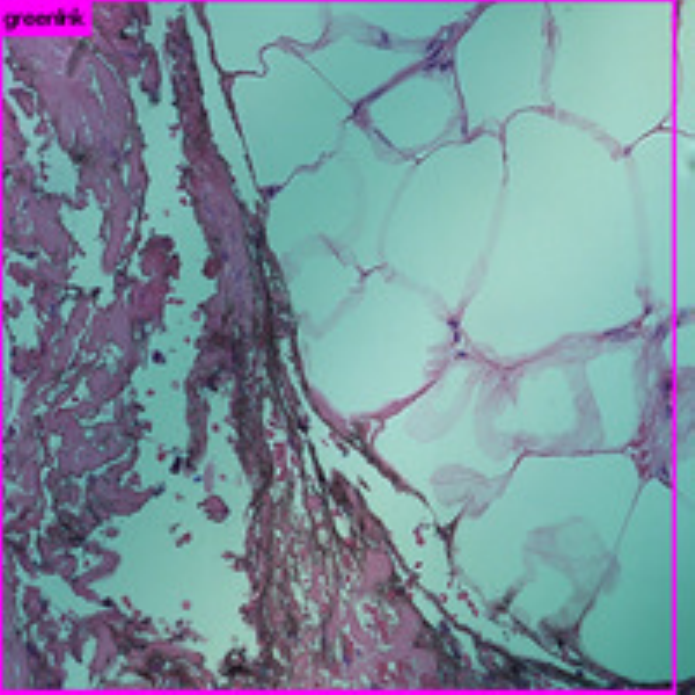}
\end{minipage}

\begin{minipage}[b]{0.32\linewidth}
\includegraphics[scale=0.4]{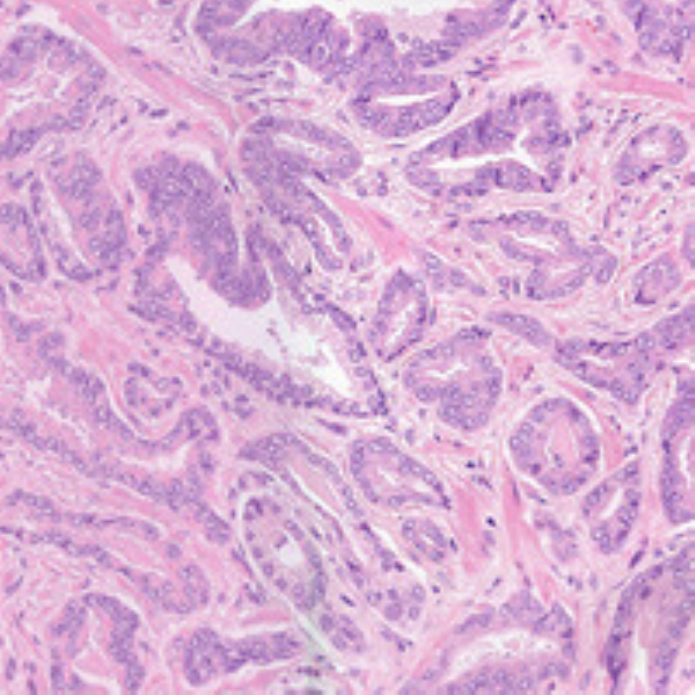}
\centerline{(a)}\medskip
\end{minipage}
\begin{minipage}[b]{0.32\linewidth}
\includegraphics[scale=0.4]{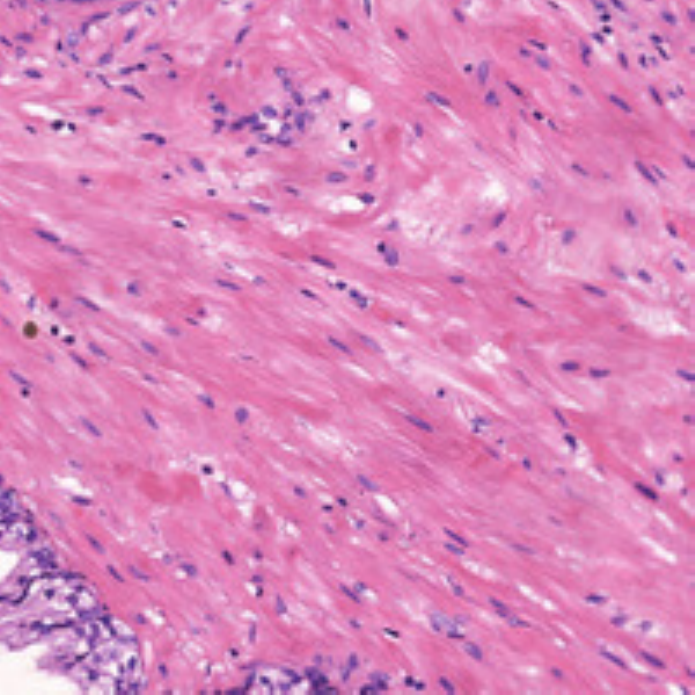}
\centerline{(b)}\medskip
\end{minipage}
\begin{minipage}[b]{0.32\linewidth}
\includegraphics[scale=0.4]{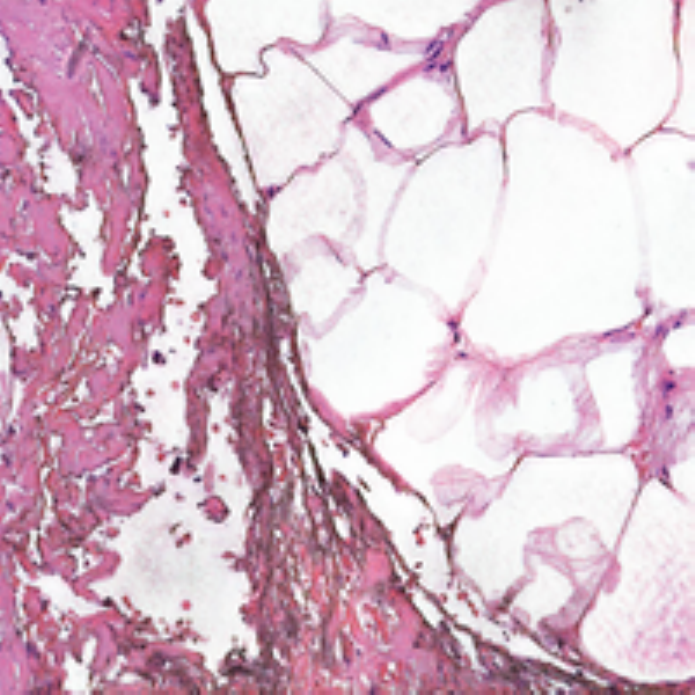}
\centerline{(c)}\medskip
\end{minipage}
\caption{{\bf Ink removal examples} \it Top: a) Identified contamination of a portion of the image, (b) and (c) identified contamination of the entire tile. Bottom: restoration results using cycleGAN with (a-b) dense  and (c) sparse domain weights. It is to be noted that only the idenified corrupted part was restored in (a). Pink and green bounding-boxes on the top row represent ink and cell cluster areas, respectively.\label{fig:bbox_restoration}}
\end{figure}
Precise localization of the ink contaminated areas within each of image tiles can substantially help in preserving underlying tissue information. That is, only the areas that are contaminated with ink are restored while uncontaminated region is kept intact (see Fig.~\ref{fig:bbox_restoration} (a), top, pink box). Secondly, we identify image tiles with clustered cells or cytoplasm (see Fig.~\ref{fig:bbox_restoration} (a-b), top, green box). We have used fast and accurate Yolov3 single-shot detector~\cite{redmon2018yolov3}. It predicts simultaneously class and bounding box coordinates using convolution filters and skip connections in multi-scale approach. We have trained darknet-53 model of Yolov3 for 2 class (ink and cluster) using bounding box annotations on 550 image tiles.
\subsection{CycleGANS for generating ink-free images}
Generative adversarial networks~\cite{Goodfellow:NIPS14} have become a powerful tool for image generation. As the original uncorrupted image is not available to us, we use CycleGAN which allows to map source domain to target domain without need for a paired image-to-image mapping. The model consists of two generator-discriminator pairs that operates cyclically for mapping image in one domain (${A}$) to another domain ($B$). While, generator $G_B$ generates image similar to domain $B$, discriminator~$D_B$ evaluates its truthfulness. The generated images are then mapped to domain $A$ back utilizing another generator-discriminator pair of domain $A$. In order to achieve cycle-consistent mapping function an $l1$ regularization was introduced in~\cite{CycleGAN2017}. Similar cyclic process is repeated in reverse direction for mapping an image in domain $B\rightarrow A$. In gist, the generator and discriminator plays a game until Nash equilibrium is achieved, i.e., the generator's distribution becomes same as the desired distribution.

Because of the variation of the underlying data distribution domain learning in histopathology data needs to be carefully defined. In order to prevent from over stretching of two distributions, we propose to utilize two different data distribution pairs based on cell or cytoplasmic mass clusters, namely: a) dense and b) sparse. We utilized $\approx 1000$ samples each for training upto 400 epochs with 100 iterations. During the restoration process, the distribution is identified by our object detector (see Section~\ref{sec:detector}) for which corresponding domain mapping is applied. In Fig.~\ref{fig:bbox_restoration} (a-b) domain learning from dense distribution is applied while sparse in case of Fig.~\ref{fig:bbox_restoration} (c) yielding a very high-quality restorations.
\section{Experiments}
%
%

Eight Formalin-Fixed Paraffin-Embedded (FFPE) H\&E (Hematoxylin and Eosin) stained prostate $75\times 50$ mm slides scanned using the Objective Imaging desktop scanner at a resolution 40X were used for this study. These slides included ink marks in four different colours (green, blue, black and red) that appeared separately or in combination. High-resolution image tiles ($1578\times 1578$ pixels) generated by the scanner were used for our ink removal experiments. For training, these image tiles were scaled to appropriate resolutions (see Table~\ref{tab:average_precision}).~However, the trained weights for both localization and image generation were applied to original image tile resolution. Restored image tiles were converted to BigTiff whole slide image format using OIWorkspaceConverter\footnote{http://www.objectiveimaging.com/download/software.php\label{footnote1}} for inspection by pathologists. Whole slide images were viewed in full detail using an standalone workspace viewer (OISwsViewer~\textsuperscript{\ref{footnote1}}). The average size of each whole slide image was around 4.5~GB.

We evaluated our sequential CNN model using an average of 549 image tiles and Yolov3 model using 100 image tiles with ground truth annotations. In order to obtain quantitative results for cycleGAN, we used CMYK color space transformation to add ink color markings (from only background inked tiles in our dataset) on 20 non-corrupted image tiles in different proportions.

\begin{table*}[t!]
	\begin{center}
		\begin{tabular}{|l|c|c|c|c|c|c|}
			\hline
			Method & time/epochs & $\bar{\text{Acc.}}$/mAP$^*$ & Size & Test load time & Train samples & Checkpoint size\\
			&mins./\# & \% & pixel & sec & \# & MB\\
			\hline\hline
			Seq. CNN & ~10 / 500 & 95 & $128\times128 $ & 0.08 &1000 (1746)& 5.2 \\
			\hline
			Yolov3~\cite{redmon2018yolov3} & ~ 120/6000 & 75$^*$ & $512\times512$ & 0.2 & 550  & 246.3 \\ 
			\hline
			CycleGAN~\cite{CycleGAN2017} &  ~1080/400 & - & $256\times256$& 0.12& 532 & 11.38 \\ 
			\hline
		\end{tabular}
	\end{center}
	\vspace{-0.5cm}
	\caption{Quantitative information for different networks used in our proposed ink removal pipeline. All timings are provided for NVIDIA Tesla P100 GPU.}
	\label{tab:average_precision}
\end{table*}

\subsection{Quantitative results}
Table~\ref{tab:average_precision} shows average accuracy ($\bar{\text{Acc.}}$) of 95\% for our sequential CNN model and a mean average precision (mAP) of 75\% for Yolov3. Our binary classifier has light weight (5.2 MB) and is computationally very fast both for training (10 mins) and at test time (0.08 s). Experimental results on cycleGAN using distributed weights (sparse-dense) showed an improvement in all the image quality metrics, notably 28.38 dB to 28.73 dB, 0.69 to 0.71, and 0.75 to 0.78, respectively for mean values of PSNR (Peak Signal to Noise Ratio), SSIM (Structure Similarity Measure) and VIF (Visual Information Fidelity). Some notable improvements for 5 simulated inked image tiles are provided in Tab.~\ref{tab:chosen}. It is worth noting that the tile \#1 has an improvement in PSNR over 4 dB while all other tiles have more than 1 dB improvement.
\begin{table}[t!]
\begin{tabular}{|c|c|c|c|c|c|c|c|}
\hline
\multirow{2}{*}{\footnotesize{Tile}}                                            & \multicolumn{2}{c|}{\footnotesize{PSNR}}                     & \multicolumn{2}{c|}{\footnotesize{SSIM}}                     & \multicolumn{2}{c|}{\footnotesize{VIF}}    \\  \cline{2-7}    
 & \footnotesize{Inked}                   & \footnotesize{Restored}              & \footnotesize{Inked}                   & \footnotesize{Restored}              &\footnotesize{Inked }                & \small{Restored}    \\   \hline
\#1 & 27.10       & 31.16        & 0.69       & 0.87        & 0.70         & 0.86           \\ \hline
\#2 & 27.43       & 29.11       & 0.68       & 0.81     & 0.75         & 0.79              \\ \hline
\#3 & 27.66       & 28.81      & 0.73       & 0.73       & 0.79         & 0.81             \\ \hline
\#4 & 27.68       & 28.52       & 0.74       & 0.78            & 0.71         & 0.80       \\ \hline
\#5 & 27.87       & 28.32         & 0.50       & 0.60  & 0.63         & 0.66               \\ \hline
\end{tabular}
\caption{{\bf Quantitative results.} \it Image quality improvement for restoration of 5 simulated inked tile images using proposed sparse-dense CycleGAN .{\label{tab:chosen}}}
\end{table}
%
\begin{figure}[t!]
\centering
%
\begin{minipage}[b]{0.45\linewidth}
\includegraphics[trim=0.25cm 0.25cm 0.5cm 0.8cm, clip=true,scale=0.23]{11.pdf}
\end{minipage}
\begin{minipage}[b]{0.45\linewidth}
\includegraphics[trim=0.25cm 0.25cm 0.5cm 0.0cm, clip=true,scale=0.27]{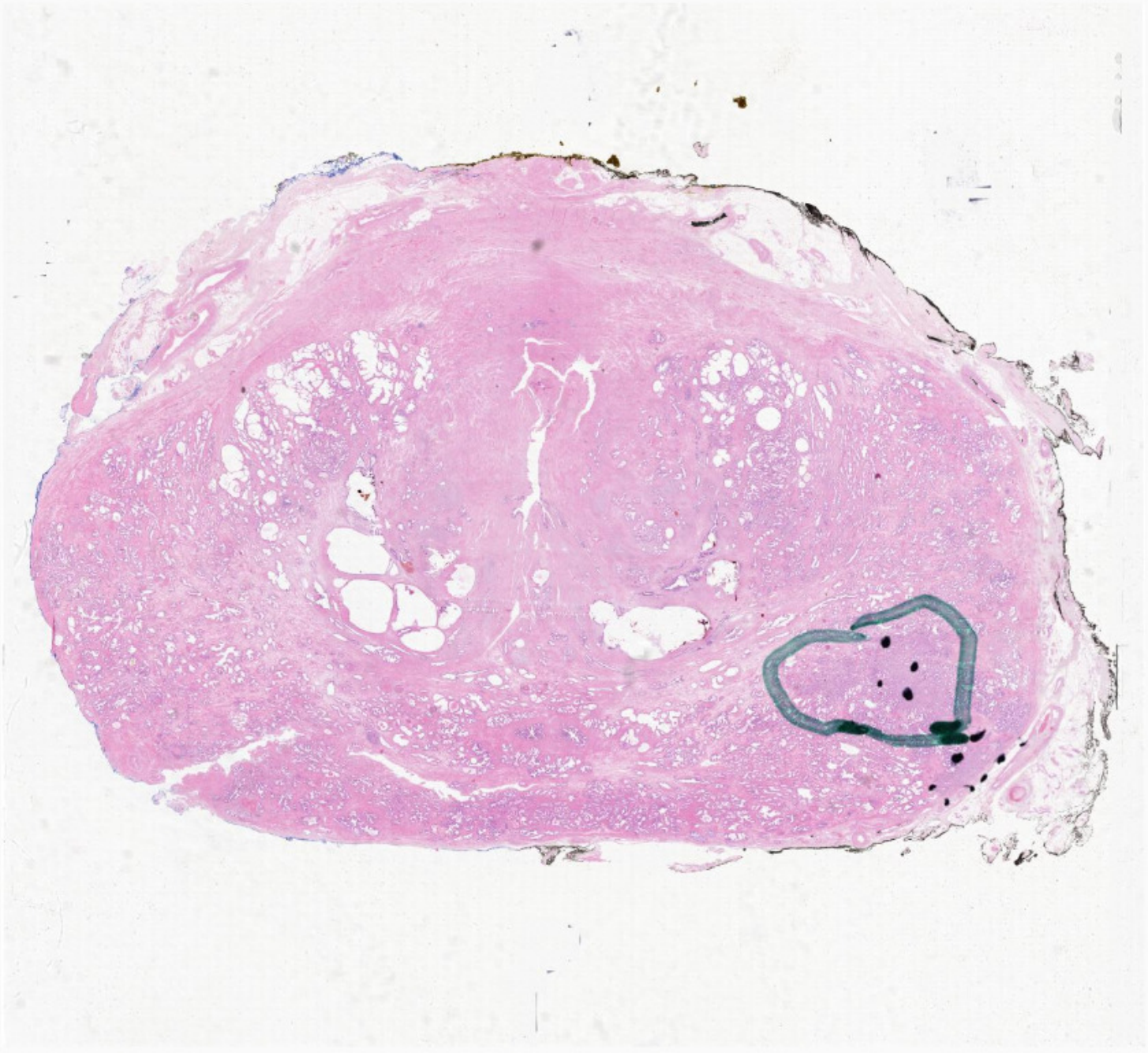}
\end{minipage}

\begin{minipage}[b]{0.45\linewidth}
\includegraphics[trim=0.25cm 0.25cm 0.0cm 0.8cm, clip=true,scale=0.17]{16.pdf}
\end{minipage}
\begin{minipage}[b]{0.45\linewidth}
\includegraphics[trim=1.25cm 0.25cm 0.5cm 0.8cm, clip=true,scale=0.21]{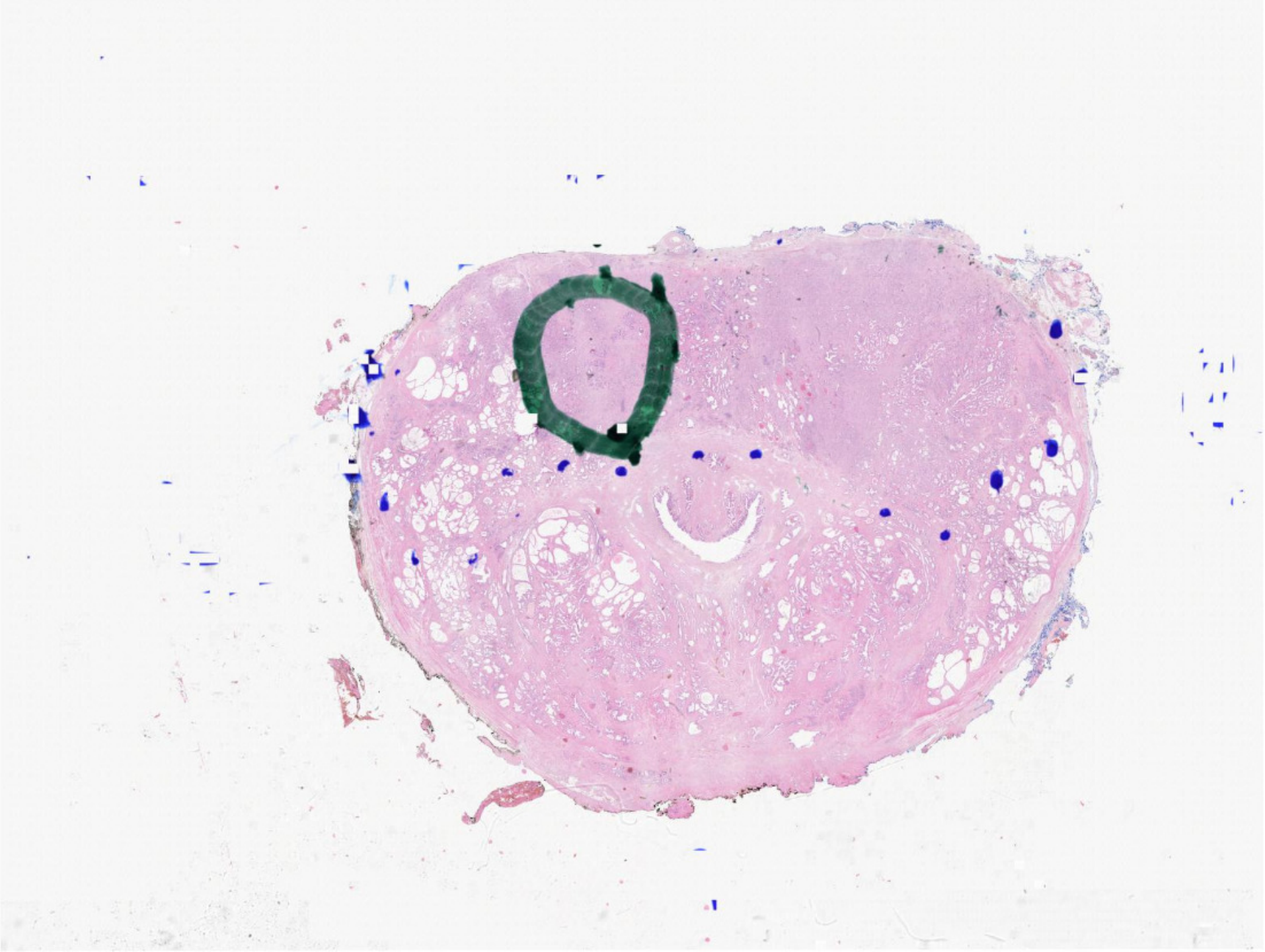}
\end{minipage}
{\caption{{\bf Background ink removal.} \it Ink removal from background of histopathology whole slide images using only CNN-based binary classifier. Left: original image corrupted with ink, right: restored image. At most times removing only background markings is sufficient.{\label{fig:bckgndInkRemoval}}}}
\end{figure}
\begin{figure}[t!]
\centering
\begin{minipage}[b]{0.45\linewidth}
\includegraphics[trim=2.5cm 0cm 3.0cm 0.1cm, clip=true,scale=0.235]{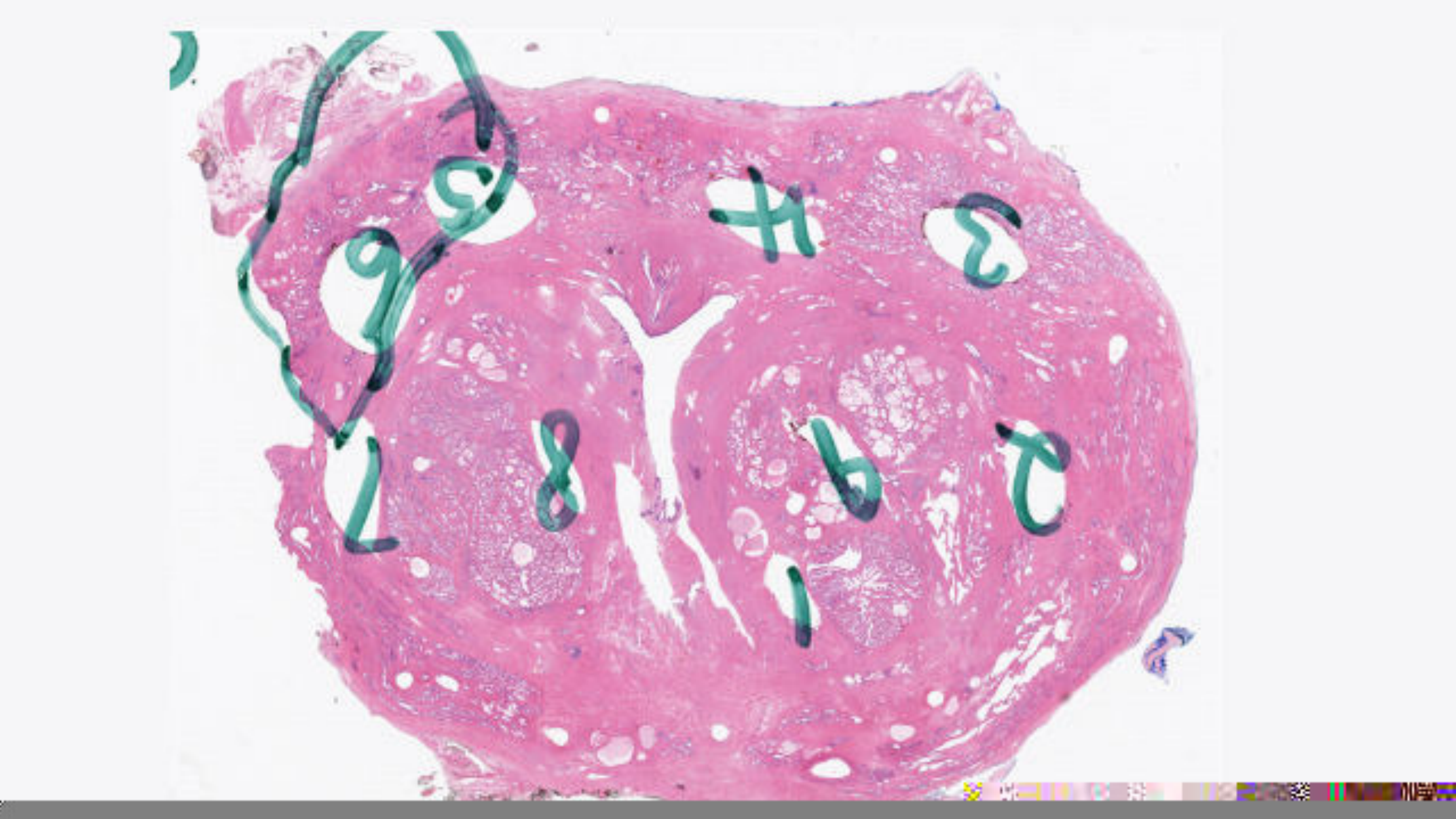}
\end{minipage}
\begin{minipage}[b]{0.45\linewidth}
\includegraphics[trim=2.5cm 0cm 3.0cm 0.1cm, clip=true,scale=0.235]{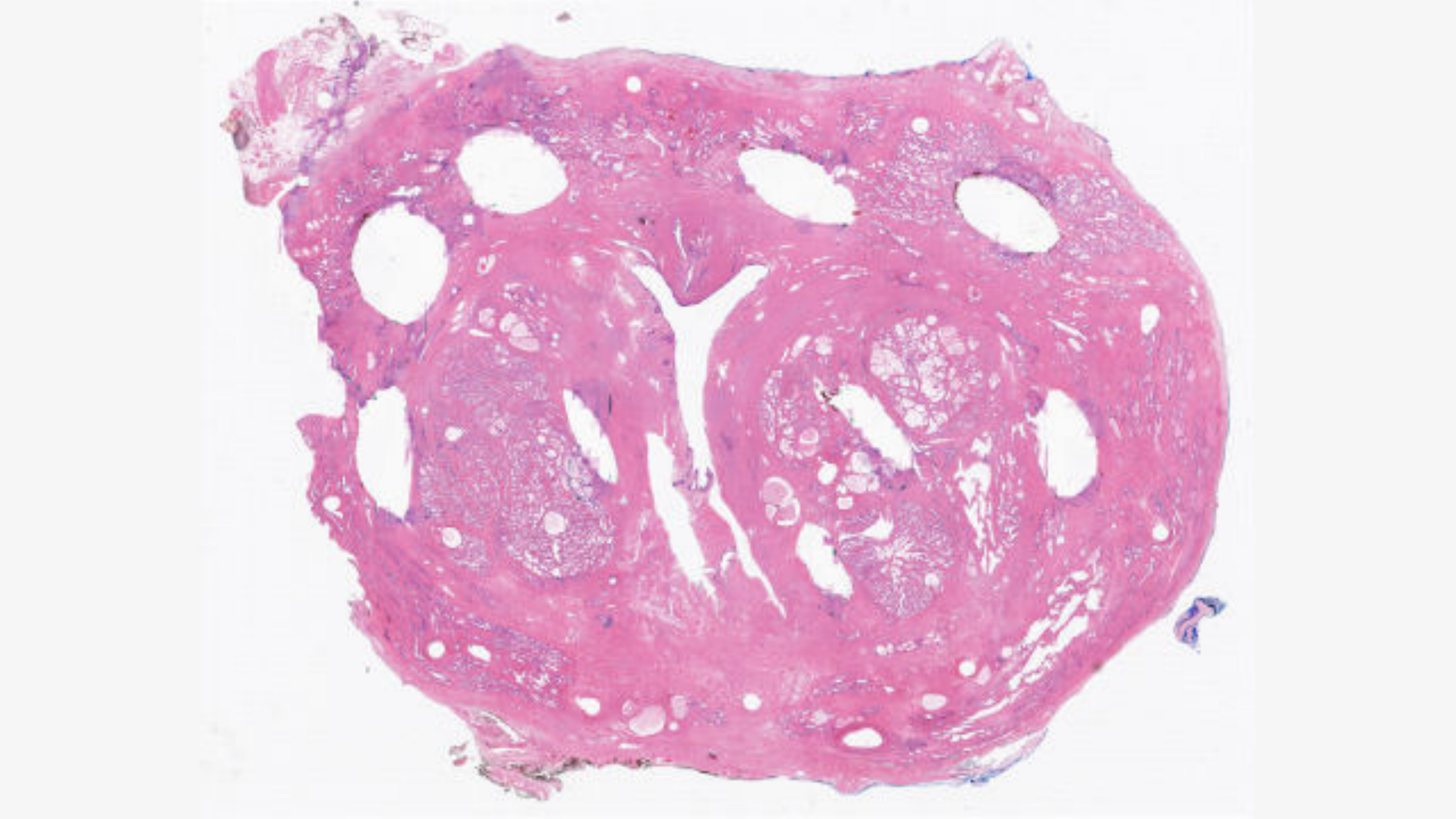}
\end{minipage}

%
{\caption{{\bf Foreground and background ink removal.} \it Ink removal from foreground and background of histopathology whole slide images using our pipeline. Left: original image corrupted with ink, right: restored image. Images have been scaled to 10\% of original image size.{\label{fig:foregndInkRemoval}}}}
\end{figure}
\begin{figure}[t!h!]
\centering
\begin{minipage}[b]{0.23\linewidth}
\includegraphics[scale=0.28]{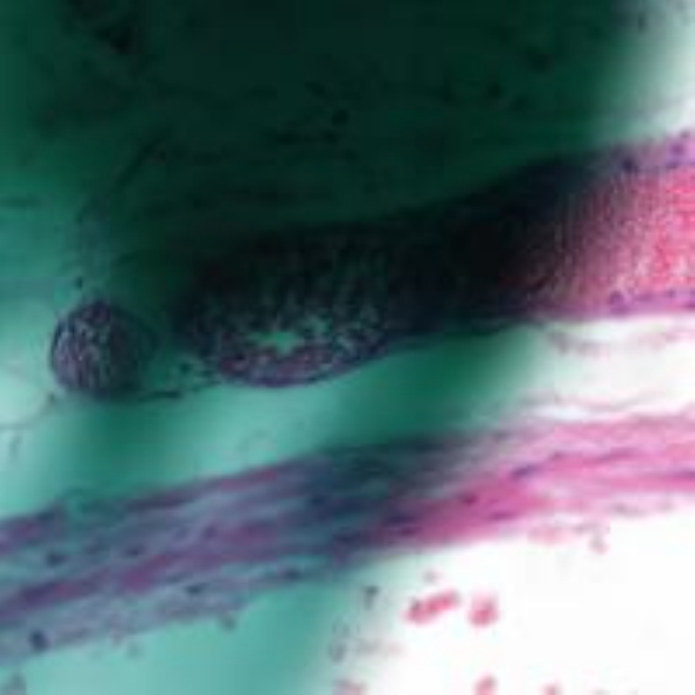}
\end{minipage}
\begin{minipage}[b]{0.23\linewidth}
\includegraphics[scale=0.28]{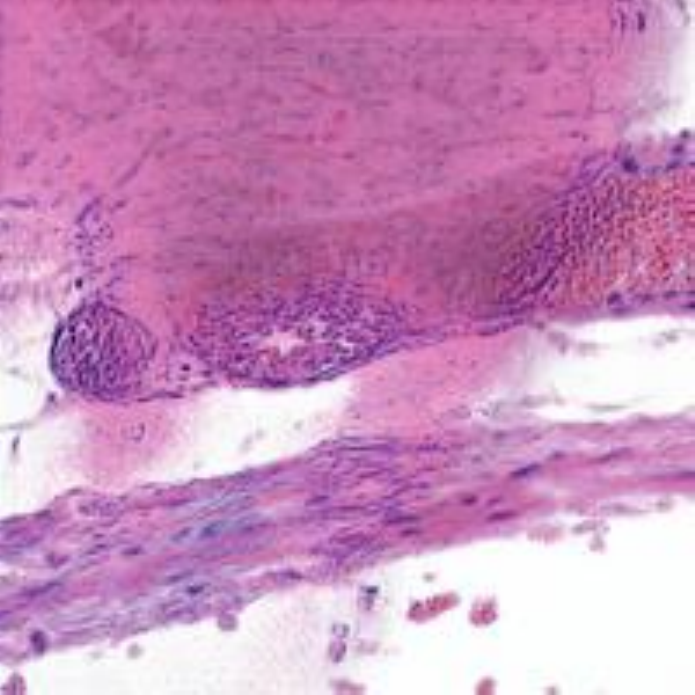}
\end{minipage}
\begin{minipage}[b]{0.23\linewidth}
\includegraphics[scale=0.28]{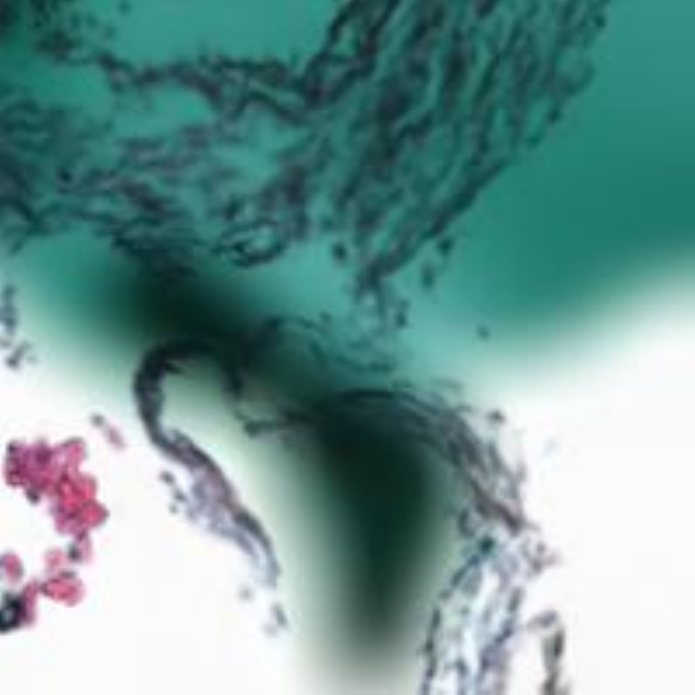}
\end{minipage}
\begin{minipage}[b]{0.23\linewidth}
\includegraphics[scale=0.28]{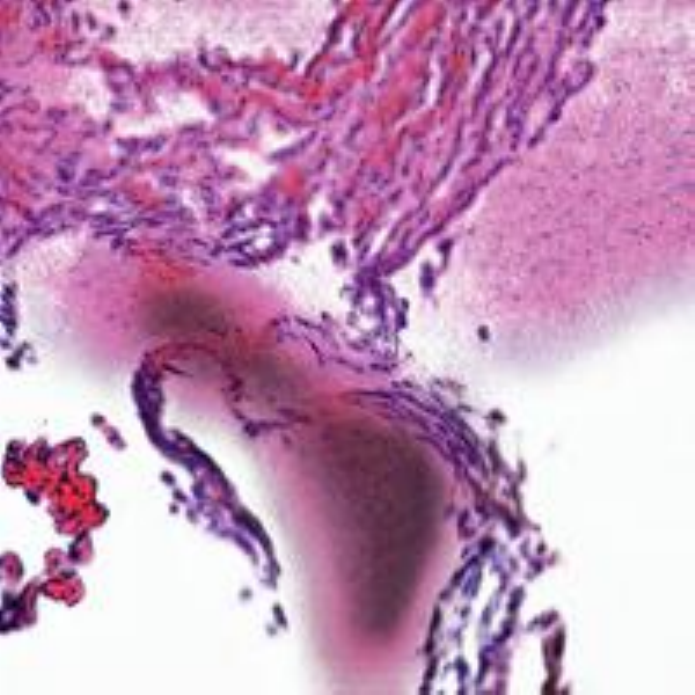}
\end{minipage}
{\caption{{\bf{Failed cases.} \it Added textures in restored image tiles.}{\label{fig:failure}}}}
\end{figure}
%
%
%
\vspace{-0.2cm}
\subsection{Qualitative results}
Visual analysis for fast and accurate ink removal from \textit{background only} utilizing solely our binary sequential CNN is shown in Fig.~\ref{fig:bckgndInkRemoval}. Other markers like dots and circles as in Fig.~\ref{fig:bckgndInkRemoval} alone do not contain any evidential information once the surrounding texts are removed. 5 out of 8 of our whole slide images had marker inks mostly on the background.~Fig.~\ref{fig:foregndInkRemoval}~(left) shows marker ink mostly on foreground pixels that also include parts of tissues. We utilize our complete pipeline process to realistically remove and restore ink contaminated regions.~ Fig.~\ref{fig:foregndInkRemoval}~(right) shows restored whole slide image which is visually coherent and marker ink has been substantially removed from both background and foreground. Fig.~\ref{fig:failure} presents some of failure results where the restoration is not perfect and addition of some texture can be clearly observed.
\vspace{-0.2cm}
\section{Conclusion}
We have proposed a content-aware and fully automatic deep learning based pipeline for efficient removal of marker inks from whole slide images without sacrificing information. We performed both quantitative and qualitative evaluations of each of the method separately which showed the efficacy of the methods. This work addresses a key issue faced by pathologists regarding usability of whole-slide images for study and research. 
%
%
%
\vspace{-0.25cm}
\section*{Acknowledgement}
\vspace{-0.25cm}
SA, NK and CV are supported by the NIHR Oxford BRC. JR is funded by EPSRC EP/M013774/1 Seebibyte.
\bibliographystyle{IEEEbib}
\bibliography{bib_histology}

\end{document}